%% file: IEEE-conference-template-062824.tex
\documentclass[conference]{IEEEtran}
\IEEEoverridecommandlockouts
\usepackage{cite}
\usepackage{amsmath,amssymb,amsfonts}
\usepackage{algorithmic}
\usepackage{graphicx}
\usepackage{textcomp}
\usepackage{subfigure}
\usepackage{xcolor}
\usepackage{multirow}
\usepackage{multicol}
\usepackage{caption}
\usepackage{algorithm}
\usepackage{algorithmic}
\def\BibTeX{{\rm B\kern-.05em{\sc i\kern-.025em b}\kern-.08em
    T\kern-.1667em\lower.7ex\hbox{E}\kern-.125emX}}
\begin{document}

\title{A Margin-Maximizing Fine-Grained \\Ensemble Method}

\author{
\IEEEauthorblockN{Jinghui Yuan*}
\IEEEauthorblockA{{\fontsize{9pt}{11pt}\selectfont\textit{School of Artificial Intelligence, OPtics and ElectroNics (iOPEN)}} \\
\textit{Northwestern Polytechnical University}\\
Shaanxi, China \\
yuanjh@mail.nwpu.edu.cn}
\thanks{* \textbf{Equal contributions.}}
\and
\IEEEauthorblockN{Hao Chen*}
\IEEEauthorblockA{{\fontsize{9pt}{11pt}\selectfont\textit{Queen Mary School Hainan}} \\
\textit{Beijing University of Posts and Telecommunications}\\
Beijing, China \\
HaoChenn.Eric@gmail.com}
\and
\IEEEauthorblockN{Renwei Luo}
\IEEEauthorblockA{{\fontsize{9pt}{11pt}\selectfont\textit{School of Artificial Intelligence, OPtics and ElectroNics (iOPEN)}} \\
\textit{Northwestern Polytechnical University}\\
Shaanxi, China \\
rweiluo@gmail.com}
\and
\IEEEauthorblockN{Feiping Nie$^\dagger$}
\IEEEauthorblockA{{\fontsize{9pt}{11pt}\selectfont\textit{School of Artificial Intelligence, OPtics and ElectroNics (iOPEN)}} \\
\textit{Northwestern Polytechnical University}\\
Shaanxi, China \\
feipingnie@gmail.com}\thanks{$\dagger$ \textbf{Corresponding\ author.}}
}

\maketitle
\input{tex/0_abstract}
\input{tex/00_keywords}
\input{tex/1_introduction}
\input{tex/2_method}
\input{tex/3_optimization}
\input{tex/4_experiment}
\input{tex/5_conclusion}

\vfill\pagebreak
\bibliographystyle{IEEEtran}
\bibliography{refs}

\end{document}

%% file: tex/0_abstract.tex
\begin{abstract}
Ensemble learning has achieved remarkable success in machine learning, but its reliance on numerous base learners limits its application in resource-constrained environments. This paper introduces an innovative “\textbf{Margin-Maximizing Fine-Grained Ensemble Method}” that achieves performance surpassing large-scale ensembles by meticulously optimizing a small number of learners and enhancing generalization capability. We propose a novel learnable confidence matrix, quantifying each classifier's confidence for each category, precisely capturing category-specific advantages of individual learners. Furthermore, we design a margin-based loss function, constructing a smooth and partially convex objective using the logsumexp technique. This approach improves optimization, eases convergence, and enables adaptive confidence allocation. Finally, we prove that the loss function is Lipschitz continuous, based on which we develop an efficient gradient optimization algorithm that simultaneously maximizes margins and dynamically adjusts learner weights. Extensive experiments demonstrate that our method outperforms traditional random forests using only one-tenth of the base learners and other state-of-the-art ensemble methods.
% 集成学习在机器学习领域取得显著成功，但其对大量基础学习器的依赖限制了在资源受限环境中的应用。本文提出一种创新的"边际最大化的细粒度集成方法"，通过精细优化少量学习器和提升泛化能力，实现了与大规模集成相当甚至更优的性能。我们提出新颖的可学习的置信度矩阵Θ，精确捕捉各学习器的类别特定优势。此外，我们设计了基于margin的损失函数，利用logsumexp技术构造平滑且部分凸的目标函数，并证明其梯度矩阵列和为零，实现自适应置信度分配。最后，我们开发高效的梯度优化算法，同步最大化margin和动态调整学习器权重。大量实验表明，我们的方法仅使用1/10的基础学习器即可超越传统随机森林和其他先进集成方法。
\end{abstract}

%% file: tex/00_keywords.tex

%% file: tex/1_introduction.tex
\section{INTRODUCTION}
\label{sec:intro}

Ensemble learning\cite{zhang2012ensemble}, a core method in machine learning, builds powerful predictive models by integrating predictions from multiple base learners. It has demonstrated exceptional performance across various domains\cite{zhou2021domain, mohammed2023comprehensive}, from computer vision\cite{haeusler2013ensemble} to natural language processing\cite{sangamnerkar2020ensemble}, bioinformatics\cite{verma2017comparative, wang2019simlin}, and financial forecasting\cite{bui2018novel}. However, as application scenarios become more complex and computational resources become constrained, traditional ensemble methods face significant challenges in resource-limited environments. Notably, the issue of model efficiency and interpretability remains unresolved, with recent methods failing to maintain or improve model performance while substantially reducing the number of base learners.

% 集成学习作为机器学习领域的核心方法之一，通过整合多个基础学习器的预测来构建强大的预测模型，在诸多领域展现出卓越的性能。从计算机视觉到自然语言处理，从生物信息学到金融预测，集成方法不断刷新各项任务的性能基准。然而，随着应用场景的复杂化和计算资源的限制，传统集成方法面临着资源受限环境中的广泛应用的重大挑战。值得注意的式，这一关乎模型的效率和可解释性的问题仍没被解决，最新的方法无法在显著减少基础学习器数量的同时保持或提升模型的性能。
Recent years have seen various approaches to optimize ensemble learning. Classical methods like Bagging\cite{ngo2022evolutionary, tuysuzouglu2020enhanced} and Boosting\cite{ying2013advance, webb2004multistrategy} enhance performance by increasing base learner diversity. Algorithms such as XGBoost\cite{chen2016xgboost, sagi2021approximating} introduced gradient boosting, further improving accuracy. Recent research, like improved random forests\cite{sun2024improved}, attempts to reduce decision tree correlations. 

While these methods have improved ensemble model performance, they still rely on numerous base learners for optimal results. Traditional ensemble methods often employ simple averaging or global performance-based weighting, overlooking differentiated performance across classification scenarios. Some studies have introduced margin\cite{liu2024nonlinear, li2021beyond, feng2018class, hu2014exploiting, li2014exploration} concepts to enhance generalization, but often lead to difficult-to-optimize problems. These limitations highlight deficiencies in current ensemble learning methods regarding fine-grained integration, efficient use of limited base learners, and generalization capabilities\cite{yuan2024achievinglesstensoroptimizationpoweredensemble}.

% 近年来，研究者们提出了多种方法来优化集成学习的性能。Bagging和Boosting等经典方法通过增加基础学习器的多样性来提升整体性能。随后，XGBoost等算法引入了梯度提升的思想，进一步提高了模型的准确性。最新的研究如改进的随机森林尝试减少决策树之间的相关性。尽管这些方法在一定程度上提高了集成模型的效果，但它们仍然依赖于大量的基础学习器来达到理想的性能。此外，传统的集成方法往往采用简单平均或基于整体性能的加权策略，忽视了基础学习器在不同分类场景和类别上的差异化表现。在提升模型泛化能力方面，虽然一些研究尝试引入margin概念，但往往导致难以优化的非平滑或非凸问题。这些限制突显了当前集成学习方法在精细化整合、高效利用有限基础学习器以及增强泛化能力方面的不足。
To address these challenges, we propose an innovative ensemble learning method that surpasses the performance of large-scale ensembles with fewer base learners through fine-grained learner optimization and margin-based generalization enhancement. Our main contributions are:

% 针对这些挑战，我们提出了一种创新的集成学习方法，通过细粒度的学习器优化和基于margin的泛化能力提升，实现了以少量基础学习器达到甚至超越大规模集成的效果。我们的主要贡献如下：
\begin{itemize}
\item Introduction of a learnable confidence matrix $\Theta$, enabling fine-grained optimization of base learners across categories. This approach captures each learner's unique strengths, enhancing the ensemble's accuracy and robustness, particularly for imbalanced classes and difficult samples.
% 1.引入可学习的置信度矩阵Θ，实现基础学习器在不同类别上的细粒度优化。这种方法精确捕捉每个学习器的独特优势，显著提升了集成模型的准确性和鲁棒性，尤其在处理类别不平衡和难分类样本时表现卓越。
\item Design of an innovative margin-based loss function. Using logsumexp techniques, we construct a smooth and partially convex loss function. This enables adaptive allocation of learner confidence under constraints, effectively enhancing model generalization and  greatly improving model robustness to unseen data.

\item Comprehensive experimental and theoretical analysis. Our method outperforms traditional random forests and other advanced ensemble methods using only $\frac{1}{10}$ of the base learners across various datasets. We explore its relationship with dropout, reveal its connection to random forests as a special case, and discuss extensions to stacking, advancing heterogeneous ensemble learning research.
% 2.	设计基于margin的创新损失函数。利用logsumexp技术构造平滑且部分凸的损失函数，并证明其梯度矩阵列和为零，实现了在约束条件下自适应分配学习器置信度，有效增强模型泛化能力。
% 3.	开发了高效的基于梯度的优化算法。利用损失函数的特殊性质，我们设计了一种能够同时最大化margin和动态调整学习器权重的优化方法，大幅提高了模型对未见数据的鲁棒性。
% 4.	通过广泛实验验证方法有效性并深入探讨理论基础。在多个数据集上，我们的方法以1/10的基础学习器数量超越传统随机森林和其他先进集成方法。同时，我们分析了该方法与dropout的关系，揭示了随机森林作为特例的理论联系，并探讨了向stacking框架扩展的可能性，为异构集成学习研究开辟新方向。
\end{itemize}

%% file: tex/2_method.tex
\section{METHODOLOGY}
\label{sec:pagestyle}
%Method要点：
%最开始简单介绍一下用到的符号（不用单独起一段）
%引入margin的定义，然后用logsumexp近似一下光滑的margin
%顺便这个近似有一个参数\alpah，tpami中要说明这个\alpha多少不影响实验结果，理论给一个上界，顺便实验说明一下
%引入这个交叉熵，最终导出这个损失函数
%第三部分是优化部分，就是求导就完事了
Assuming we now have $k$ classifiers, denoted as $G_1,...,G_k$, respectively. We aim to integrate these $k$ base classifiers at a fine-grained level by learning $\Theta$, where $\Theta_{ij}$ represents the confidence of the i-th class in assigning the sample to the j-th classifier. Assuming the dataset is $\{x_1,...,x_n\}$, $Y_i$ is the label of $x_i$ and $Y_i$ is a c-dimensional one-hot encoded column vector. $G_n(x_i)$ is also one-hot encoded column vector like $Y_i$. Let $g_i = [G_1(x_i),...,G_k(x_i)]^T$, with $g_i \in \mathbb{R}^{k \times c}$ and $(g_{i})_{pq}$ representing the element on p-th row and q-th column. Given $\Theta$, the prediction of the ensemble classifier can be obtained using the following formula.
\begin{equation}
    \hat{y}_i= \mathop{\arg\max}\limits_{j = 1...c}\ \mathcal{S}[\left(I\odot\Theta g_i\right)\mathbf{1}]_j
\end{equation}

Here, $I$ is the identity matrix, $\mathcal{S}[\cdot]$ is the softmax operator, which applies the softmax operation to a vector, $\mathbf{1}$ is a c-dimensional column vector filled with $1$, and $\odot$ denotes the Hadamard product, representing element-wise multiplication of matrices.

To enhance the generalization ability of the ensemble model, we aim to learn $\Theta$ such that the margin is maximized. A reasonable definition of the margin is as follows.
\begin{equation}
\label{eq4}
Y_i^T(Y_i\odot\mathcal{S}[\left(I\odot\Theta g_i\right)\mathbf{1}])-max_2[\mathcal{S}[\left(I\odot\Theta g_i\right)\mathbf{1}]]
\end{equation}

This margin is reasonable because it seeks to maximize the difference between the probability at the correct position and the probability at the second highest position. Here, $\max_2[\cdot]$ refers to the second highest value in the input vector. However, the problem lies in the fact that the $max_2$ function is non-smooth and non-convex, which poses significant challenges for optimization. We use a smooth convex function, the logsumexp function, to replace $\max_2$. Under the following assumption 
\begin{equation}
\label{Eq4}
max[\mathcal{S}[\left(I\odot\Theta g_i\right)\mathbf{1}]] = \| Y_i \odot \mathcal{S}[\left(I\odot\Theta g_i\right)\mathbf{1}] \|
\end{equation}
the $max_2$ function can be smoothly approximated like the following formula by using the logsumexp function\cite{blanchard2021accurately}. We will justify this assumption later. Here, $\alpha$ is a constant, and as long as $\alpha$ is relatively large, it does not affect the result\cite{calafiore2020universal}.
\begin{equation}
\begin{aligned}
&max_2[\mathcal{S}[\left(I\odot\Theta g_i\right)\mathbf{1}]]=\\
&\frac{1}{\alpha}\log\left(\sum_{j=1}^{c}e^{\alpha (\mathcal{S}[\left(I\odot\Theta g_i\right)\mathbf{1}]-Y_i\odot\mathcal{S}[\left(I\odot\Theta g_i\right)\mathbf{1}])_j}\right)
\end{aligned}
\end{equation}

Based on this, the margin can be smoothly represented by the following formula.
\begin{equation}
\begin{aligned}
\label{eq6}
&\mathcal{M}=\sum_{i=1}^n
 Y_i^T(Y_i\odot\mathcal{S}[\left(I\odot\Theta g_i\right)\mathbf{1}])\\
&-\sum_{i=1}^n\frac{1}{\alpha}\log\left(\sum_{j=1}^{c}e^{\alpha (\mathcal{S}[\left(I\odot\Theta g_i\right)\mathbf{1}]-Y_i\odot\mathcal{S}[\left(I\odot\Theta g_i\right)\mathbf{1}])_j}\right)
\end{aligned}
\end{equation}

To ensure that Eq.\eqref{Eq4} is reasonable, we need another component in the loss function to ensure high accuracy. That is, we need to make the ensemble learning as accurate as possible. We can use the cross-entropy loss function to guarantee this, i.e., the second part is following.
\begin{equation}
\mathcal{C}=-\sum_{i=1}^n(Y_i^T(Y_i\odot \log(\mathcal{S}[\left(I\odot\Theta g_i\right)\mathbf{1}])))
\end{equation}

Therefore, the final loss function can be expressed as the weighted sum of the two parts, given by the following formula. Our goal is to minimize the loss function $\mathcal{L}$.
\begin{equation}
    \begin{aligned}
&\mathcal{L}=\mathcal{C}-\gamma\mathcal{M}\\
&=\sum_{i=1}^n-\left(Y_i^T\log(\mathcal{S}[\left(I\odot\Theta g_i\right)\mathbf{1}])+\gamma
Y_i^T\mathcal{S}[\left(I\odot\Theta g_i\right)\mathbf{1}]\right)\\
&+\gamma  \sum_{i=1}^n\frac{1}{\alpha}\log\left(\sum_{j=1}^{c}e^{\alpha (\mathcal{S}[\left(I\odot\Theta g_i\right)\mathbf{1}]-Y_i\odot\mathcal{S}[\left(I\odot\Theta g_i\right)\mathbf{1}])_j}\right)   
    \end{aligned}
\end{equation}

It is worth mentioning that even if assumption in Eq.\eqref{Eq4} does not hold, it does not affect the validity of our loss function $\mathcal{L}$. Because even if it does not hold, the loss function introduced by $\mathcal{M}$ becomes the following.
\begin{equation}
\mathcal{M}_i=Y_i^T(Y_i\odot\mathcal{S}[\left(I\odot\Theta g_i\right)\mathbf{1}])-max(\mathcal{S}[\left(I\odot\Theta g_i\right)\mathbf{1}])
\end{equation}

This means that the optimization direction still progresses towards increasing $Y_i^T(Y_i\odot\mathcal{S}[\left(I\odot\Theta g_i\right)\mathbf{1}])$ and decreasing $max(\mathcal{S}[\left(I\odot\Theta g_i\right)\mathbf{1}])$. Once $Y_i^T(Y_i\odot\mathcal{S}[\left(I\odot\Theta g_i\right)\mathbf{1}])$  exceeds $max(\mathcal{S}[\left(I\odot\Theta g_i\right)\mathbf{1}])$, assumption in Eq.\eqref{Eq4} holds, and it then transforms into Eq.\eqref{eq6}.

%% file: tex/3_optimization.tex
\section{OPTIMIZATION ALGORITHM}
\label{sec:majhead}
% 为了充分发挥细粒度集成方法的潜力，优化过程至关重要。深入分析损失函数的数学特性不仅为算法的收敛性提供理论基础，还能指导更高效的优化策略的设计。我们将探讨损失函数的局部凸性和梯度特性，这为提升模型性能和计算效率奠定了基础。
In this section, we design the optimization algorithm. First, we prove that the loss function is convex with respect to $\mathcal{S}[\left(I\odot\Theta g_i\right)\mathbf{1}]$. Second, we demonstrate that the loss function is Lipschitz continuous and provide the Lipschitz constant, which makes our model robust and easier to optimize. Finally, we present the optimization method we used.

\textbf{Theorem 1} The loss function $\mathcal{L}$ is a convex function with respect to $\mathcal{S}[\left(I\odot\Theta g_i\right)\mathbf{1}]$

\textbf{Proof} : First, it is evident that $Y_i\odot\mathcal{S}[\left(I\odot\Theta g_i\right)\mathbf{1}]$ is a linear transformation of $\mathcal{S}[\left(I\odot\Theta g_i\right)\mathbf{1}]$.
% 利用凸函数的性质，我们知道对于凸函数f(x)，f(Ax + b)也是凸函数。结合log-sum-exp函数的已知凸性，以及其他项的明显凸性，我们可以得出损失函数L关于S(Θg_i)的凸性结论。
Leveraging the properties of convex functions, we know that for a convex function $f(x)$, $f(Ax + b)$ is also convex\cite{boyd2004convex}. Combining this with the known convexity of the logsumexp function and the evident convexity of other terms\cite{xi2020log}, we can conclude the convexity of the loss function $\mathcal{L}$ with respect to $\mathcal{S}[\left(I\odot\Theta g_i\right)\mathbf{1}]$.

A Lipschitz continuous\cite{zuhlke2024adversarial} loss function can make a machine learning model more robust\cite{weng2018evaluating}. We will prove that our loss function is Lipschitz continuous, and the Lipschitz constant $L$ is less than $\sqrt{ck} \cdot (1+\gamma+\frac{\gamma}{c} e^{\alpha})$, which facilitates easier optimization.

\textbf{Theorem 2} The loss function $\mathcal{L}$ is Lipschitz continuous $\|\mathcal{L}(\Theta)-\mathcal{L}(\tilde{\Theta})\|\le L\|\Theta-\tilde{\Theta}\|_F$. Moreover, the Lipschitz constant $L$ is less than $\sqrt{ck} \cdot (1+\gamma+\frac{\gamma}{c} e^{\alpha})$.

\textbf{Proof} : Similar to \cite{yuan2024achievinglesstensoroptimizationpoweredensemble}, the derivative of the loss function with respect to the variable $\Theta$ is following. Without loss of generality, we analyse the case with one data belongs to the m-th class, which corresponds to the m-th row of $Y$ being 1 and all other rows being 0. Let $g$ represent the result matrix, specifically the $g_i$ mentioned earlier.

% 首先，我们计算softmax函数S[(I⊙Θg_n)1]关于θ_kl的偏导数，这是后续所有计算的基础：
First, we calculate the partial derivative of the softmax function $\mathcal{S}[(I\odot\Theta g)\mathbf{1}]$ with respect to $\Theta_{kl}$, which is the basis for all subsequent calculations:
\begin{equation}
\label{eqbase}
    \begin{aligned}
\frac{\partial \mathcal{S}[(I\odot\Theta g)\mathbf{1}]_j}{\partial \Theta_{kl}}
= \frac{\big(\delta_{kj}g_{lj}-g_{lk}\mathcal{S}[(I\odot \Theta g)1]_{k}\big)}{(\mathcal{S}[(I\odot \Theta g)1]_{j})^{-1}}
    \end{aligned}
\end{equation}
% 其中，δ_jk是克罗内克delta函数。
where $\delta_{kj}$ is the Kronecker delta function\cite{kozen2007indefinite} and it equals to 1 only if $k=j$, else 0.
% 不失一般性，我们假设只有一个数据，属于第 m 类。相当于Y的第m行为1，其他行均为0。因此，不需要求和，损失函数可以表示为：

Therefore, there is no need for summation, and the loss function can be divided into three parts as followed:
\begin{equation}
    \begin{aligned}
&\mathcal{L}=\mathcal{L}_1+\mathcal{L}_2+\mathcal{L}_3\\
=&-Y^T\log(\mathcal{S}[\left(I\odot\Theta g\right)\mathbf{1}])-\gamma
Y^T\mathcal{S}[\left(I\odot\Theta g\right)\mathbf{1}]\\
+&\gamma \frac{1}{\alpha}\log\left(\sum_{j=1}^{c}e^{\alpha (\mathcal{S}[\left(I\odot\Theta g\right)\mathbf{1}]-Y\odot\mathcal{S}[\left(I\odot\Theta g\right)\mathbf{1}])_j}\right)   
    \end{aligned}
\end{equation}

According to the chain rule and the Eq.\eqref{eqbase}, the first term $\mathcal{L}_1=-Y^T\log(\mathcal{S}[\left(I\odot\Theta g\right)\mathbf{1}])$ with respect to $\Theta_{kl}$ is easily known to be
\begin{equation}
\begin{aligned}
\frac{\partial \mathcal{L}_1}{\partial \Theta_{kl}} = &-\big(\delta_{km}g_{lm}-g_{lk}\mathcal{S}[(I\odot \Theta g)1]_{k}\big)
\end{aligned}
\end{equation}
Similarly, it is straightforward to compute
\begin{equation}
\begin{aligned}
\frac{\partial \mathcal{L}_2}{\partial \Theta_{kl}} =-\gamma \mathcal{S}[(I\odot \Theta g)1]_{m}\big(\delta_{km}g_{lm}-g_{lk}\mathcal{S}[(I\odot \Theta g)1]_{k}\big)
\end{aligned}
\end{equation}
It is easy to verify that the derivative of the third term can be expressed in the following form, where $\tilde{\mathcal{S}}_j=\mathcal{S}[\left (I\odot\Theta g\right)\mathbf{1}]_j$ and $\sum_{j=1}^c\mathcal{I}(j\ne m)$ means compute the summary when $j\ne m$.
\begin{equation}
    \begin{aligned}
&\frac{\partial \mathcal{L}_3 }{\partial \Theta_{kl}}= 
\frac{\sum_{j=1}^c\mathcal{I}(j\ne m)e^{\alpha\tilde{\mathcal{S}}_j}\tilde{\mathcal{S}}_j\big(\delta_{kj}g_{lj}-g_{lk}\tilde{\mathcal{S}}_k\big)}{\gamma^{-1}(\sum_{j=1}^ce^{\alpha\tilde{\mathcal{S}}_j}-e^{\alpha\tilde{\mathcal{S}}_m}+1)}
    \end{aligned}
\end{equation}

% \begin{equation}
%     \begin{aligned}
% &\frac{\partial \mathcal{L}_3 }{\partial \Theta_{kl}}= \frac{\gamma\delta_{pl}}{\sum_{j=1}^ce^{\alpha\mathcal{S}[\left (I\odot\Theta g\right)\mathbf{1}]_j}-e^{\alpha\mathcal{S}[\left (I\odot\Theta g\right)\mathbf{1}]_m}+1} \cdot\\
% &[\sum_{j=1}^ce^{\alpha\mathcal{S}[\left (I\odot\Theta g\right)\mathbf{1}]_j}\mathcal{S}^2[\left (I\odot\Theta g\right)\mathbf{1}]_j\big(\delta_{kj}g_{pj}-g_{pk}\mathcal{S}[(I\odot \Theta g)1]_{k}\big)\\
% &-e^{\alpha\mathcal{S}[\left (I\odot\Theta g\right)\mathbf{1}]_m}\mathcal{S}^2[\left (I\odot\Theta g\right)\mathbf{1}]_m\big(\delta_{km}g_{pm}-g_{pk}\mathcal{S}[(I\odot \Theta g)1]_{k}\big)]
%     \end{aligned}
% \end{equation}

% 在得到损失函数L关于θ_ij的完整梯度表达式后,我们的下一个目标是证明这个梯度是有界的。这一步对于后续建立Lipschitz连续性至关重要,进而为我们应用高效的优化算法奠定基础。
After obtaining the complete gradient expression of the loss function with respect to $\Theta_{kl}$, our next objective is to prove that this gradient is bounded. This step is crucial for subsequently establishing Lipschitz continuity, which in turn lays the foundation for applying efficient optimization algorithms.

To prove the boundedness of the gradient, we need to carefully analyze each component of the gradient expression. First, we observe that the output of the softmax function, $\mathcal{S}[\left (I\odot\Theta g\right)\mathbf{1}]_j$, is always within the interval $[0,1]$. Similarly, $g_{pk}$ and $\delta_{km}$ are both either 0 or 1. Hence, $|\frac{\partial \mathcal{L}_1}{\partial \Theta_{kl}}|\le 1$ and $|\frac{\partial \mathcal{L}_2}{\partial \Theta_{kl}}|\le \gamma$.

As for the third term, since we have $\mathcal{S}[\left (I\odot\Theta g\right)\mathbf{1}]_j \leq 1$ and $|\delta_{kj}g_{lj}-g_{lk}\mathcal{S}[(I\odot \Theta g)1]_{k}| \leq 1$, and the fact that 
\begin{equation}
|\sum_{j=1}^ce^{\alpha\mathcal{S}[\left (I\odot\Theta g\right)\mathbf{1}]_j}-e^{\alpha\mathcal{S}[\left (I\odot\Theta g\right)\mathbf{1}]_m}+1| \geq c
\end{equation}
holds since the exponent is positive, we obtain the boundary as followed $|\frac{\partial \mathcal{L}_3 }{\partial \Theta_{kl}}| \leq \frac{\gamma }{c} e^{\alpha}$. Combining these results, we can deduce an overall upper bound for the gradient: $|\frac{\partial \mathcal{L}}{\partial \Theta_{kl}}|\le 1+\gamma+\frac{\gamma}{c} e^{\alpha}.$ By the Mean Value Theorem, for any two points $\Theta$ and $\tilde{\Theta}$, there exists an intermediate point $\xi$ such that $\mathcal{L}(\Theta) - \mathcal{L}(\tilde{\Theta}) = \nabla \mathcal{L}(\xi) \cdot (\Theta - \tilde{\Theta})$ and by the Cauchy-Schwarz Inequality we obtain:
% 综合这些结果,我们可以得出梯度的一个总体上界:
% 对于任意两点 $\Theta$ 和 $\tilde{\Theta}$，存在一个中间点 $\xi$，使得：% Cauchy-Schwarz
\begin{equation}
|\mathcal{L}(\Theta) - \mathcal{L}(\tilde{\Theta})| \leq \|\nabla \mathcal{L}(\xi)\|_F \cdot \|\Theta - \tilde{\Theta}\|_F
\end{equation}

% 利用给定的偏导数界计算梯度的 Frobenius 范数平方
Using the given bound on partial derivatives $|\frac{\partial \mathcal{L}}{\partial \Theta_{kl}}| \le 1+\gamma+ \frac{\gamma}{c} e^{\alpha}$, we can calculate the square of the Frobenius norm of the gradient:
\begin{equation}
\|\nabla \mathcal{L}(\xi)\|_F^2 = \sum_{k,l} (\frac{\partial \mathcal{L}}{\partial \Theta_{kl}})^2 \leq ck(1+\gamma+\frac{\gamma}{c} e^{\alpha})^2.
\end{equation}

% Lipschitz 常数
Finally we derive the Lipschitz constant $L = \sqrt{ck} \cdot (1+\gamma+\frac{\gamma}{c} e^{\alpha})$ and have proved the loss function $\mathcal{L}$ is Lipschitz continuous with a Lipschitz constant $L \leq \sqrt{ck} \cdot (1+\gamma+\frac{\gamma}{c} e^{\alpha})$. Lipschitz continuity makes our optimization more robust. We employ gradient descent\cite{amari1993backpropagation} to optimize the loss function.

\begin{algorithm}
    \caption{Gradient Descent}
    \label{Algorithm 1} 
    \begin{algorithmic}[1]
        \REQUIRE Matrix $G$
        \ENSURE $\Theta^*$
        \STATE Initialize $\Theta_0$ randomly.
        \WHILE{not converged}
            \STATE Randomly select a batch of columns $g_i$ from $G$.
            \STATE Compute $\nabla_{\Theta}\mathcal{L}(\Theta_p)$ by Eq.(11)(12)(13).
            \STATE Update $\Theta$ by $\Theta_{p+1} = \Theta_p - \beta\nabla_{\Theta}\mathcal{L}(\Theta_p)$.
        \ENDWHILE
    \end{algorithmic}
\end{algorithm}

%% file: tex/4_experiment.tex
\section{EXPERIMENT}
\label{sec:typestyle}

\begin{table*}
\caption{The accuracy metrics of our algorithm on various datasets}
\label{table:results}
\centering
\small
\setlength{\tabcolsep}{4pt}
\begin{tabular}{||c||c||ccccccccc||}
\hline
Data & Categories & \ OUR \ &\  RF50 \ & \ RF100 \ &\  SVM \ & XGBoost & LightGBM & \# Object & \# Attribute & \# Class  \\
\hline\hline
\multirow{2}{*}{TOY}
& Train & \ \underline{98.56}\ &\ 98.44	\ &	\ 98.38	\ &\ 98.00\ &	\textbf{98.69}	&	98.56 & 1600 & \multirow{2}{*}{2} & \multirow{2}{*}{2}  \\
& Test & \ \textbf{98.75}\ 	&\ 98.00\ &\ 98.50\ &\ 98.50\ 	&	98.00	&	97.75 & 400 &  &   \\
\hline
\multirow{2}{*}{BASEHOCK}
& Train & \ 95.80\ 	&	\ 92.16	\ &\ 	92.35\ 	&	\ \textbf{98.81}\ 	&	\underline{95.92}	&	95.17 & 1594 & \multirow{2}{*}{4862} & \multirow{2}{*}{2}  \\
& Test & \ \underline{93.73}\ 	&	\ 88.72\ 	&\ 	88.72\ 	&	\ \textbf{95.24}\ 	&	90.00	&	93.23 & 399 &  &   \\
\hline
\multirow{2}{*}{breast uni}
& Train &\  \underline{99.82}\ 	&\ 	\textbf{100}\ 	&\ 	\textbf{100}	\ &	\ 64.94	\ &	99.28	&	97.5 & 559 & \multirow{2}{*}{10} & \multirow{2}{*}{2}  \\
& Test & \ \underline{95.71}\ 	&\ 	95.00\ 	&\ 	95.00\ 	&\ 	67.86\ 	&	95.71	&	\textbf{96.43} & 140 &  &   \\
\hline
\multirow{2}{*}{chess}
& Train &\  \textbf{99.77}\ 	&\ 	99.57\ 	&\ 	\underline{99.61}\ 	&\ 	94.17\ 	&	99.50	&	98.90 & 2557 & \multirow{2}{*}{36} & \multirow{2}{*}{2}  \\
& Test &\  \textbf{99.06}\ 	&\ 	\underline{98.44}\ 	&\ 	98.44\ 	&\ 	92.5\ 	&	98.44	&	97.19 & 639 &  &   \\
\hline
\multirow{2}{*}{iris}
& Train &\  \textbf{100}\ 	&\ 	100\ 	&\ 	\underline{100}\ 	&\ 	95.50\ 	&	98.33	&	98.33 & 120 & \multirow{2}{*}{4} & \multirow{2}{*}{3}  \\
& Test & \ \textbf{96.67}\ 	&	\ 93.33	\ &	\ 93.33\ 	&\ 	96.67\ 	&	\underline{96.67}	&	96.67 & 30 &  &  \\
\hline
\multirow{2}{*}{jaffe}
& Train & \ \textbf{100}\  & \ 100\  &\  \underline{100}\  &\  100\  & 100 & 100 & 170 & \multirow{2}{*}{1024} & \multirow{2}{*}{10}  \\
& Test & \ \underline{97.67}\ 	&\ 	95.35\ 	&\ 	97.67\ 	&	\ \textbf{100}\ 	&	93.02	&	93.02 & 43 &  &  \\
\hline
\multirow{2}{*}{pathbased}
& Train &\ \textbf{100}\ &\ 	100\ 	&\ 	\underline{100}\ 	&\ 	99.17\ 	&	100	&	99.58 & 240 & \multirow{2}{*}{2} & \multirow{2}{*}{3}  \\
& Test &\ \textbf{100}\ &	\ 100\ 	&\ 	\underline{100}	\ &	\ 96.67	\ &	95.00	&	98.33 & 60 &  &   \\
%  \hline
% \multirow{2}{*}{PCMAC}
% & Train & \ 89.77\ 	&	\ 87.45\ 	&\ 	87.45	\ &	\ \textbf{96.52}\ 	&	\underline{90.09}	&	88.8 & 1554 & \multirow{2}{*}{3289} & \multirow{2}{*}{2}  \\
% & Test & \ \underline{87.15}\ 	&\ 	85.35\ 	&	\ 85.35\ 	&\ 	\textbf{91.77}\ 	&	86.89	&	87.15 & 389 &  &   \\
\hline
\multirow{2}{*}{RELATHE}
& Train &\  91.67\ 	&\ 	88.61\ 	&\ 	88.52\ 	&\ 	\textbf{95}	\ &	\underline{94.39}	&	89.66 & 1142 & \multirow{2}{*}{4322} & \multirow{2}{*}{2}  \\
& Test & \ \underline{86.71}\ 	&\ 	82.17\ 	&\ 	81.47\ 	&\ 	\textbf{90.56}\ 	&	85.31	&	85.66 & 285 &  &   \\
\hline
\multirow{2}{*}{wine}
& Train &\  \textbf{100}\  &\  100\  & \ \underline{100}\  & \ 71.83 \ & 100 & 100 & 142 & \multirow{2}{*}{13} & \multirow{2}{*}{3}  \\
& Test &\  \textbf{97.22}\ 	&\ 	97.22\ 	&\ 	\underline{97.22}\ 	&\ 	69.44\ 	&	91.67	&	91.67 & 36 &  &   \\
\hline
\end{tabular}
\end{table*}
\begin{figure}[h]
        \centering
        \subfigure[DATA]
	{
		\includegraphics[width=0.22\textwidth]{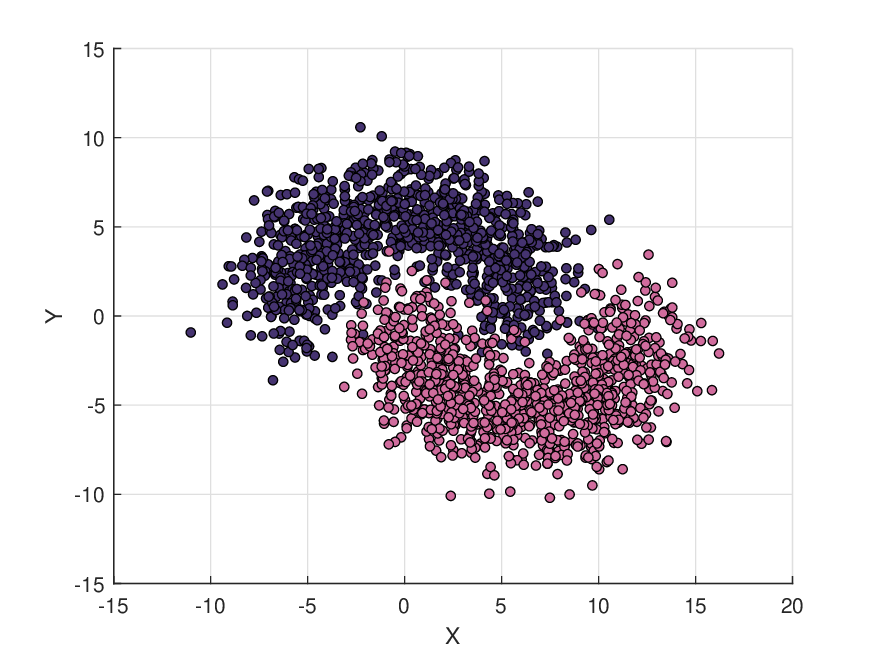}
	}
         \subfigure[OUR]
	{
		\includegraphics[width=0.22\textwidth]{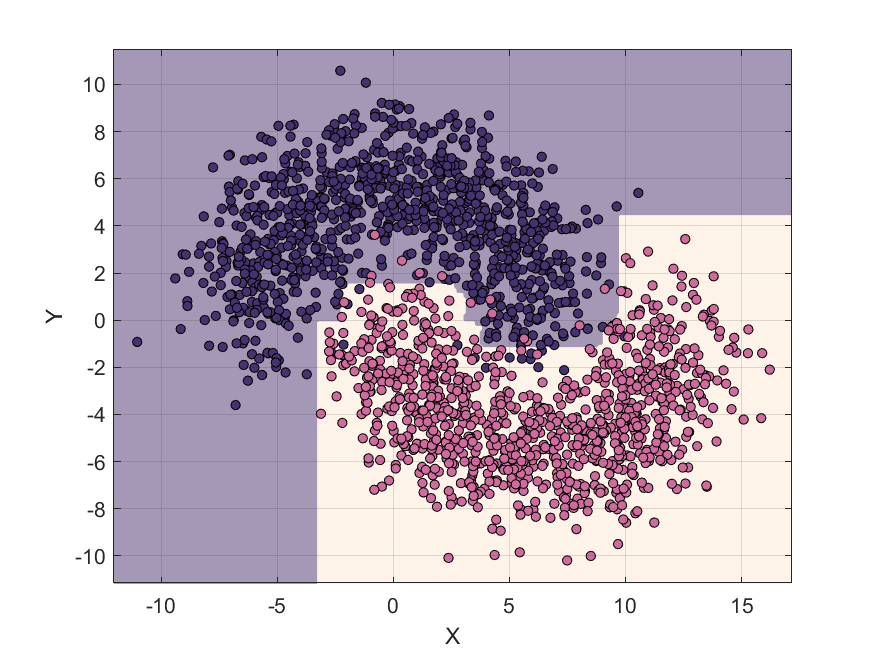}
	}
	\subfigure[RF100]
	{
		\includegraphics[width=0.22\textwidth]{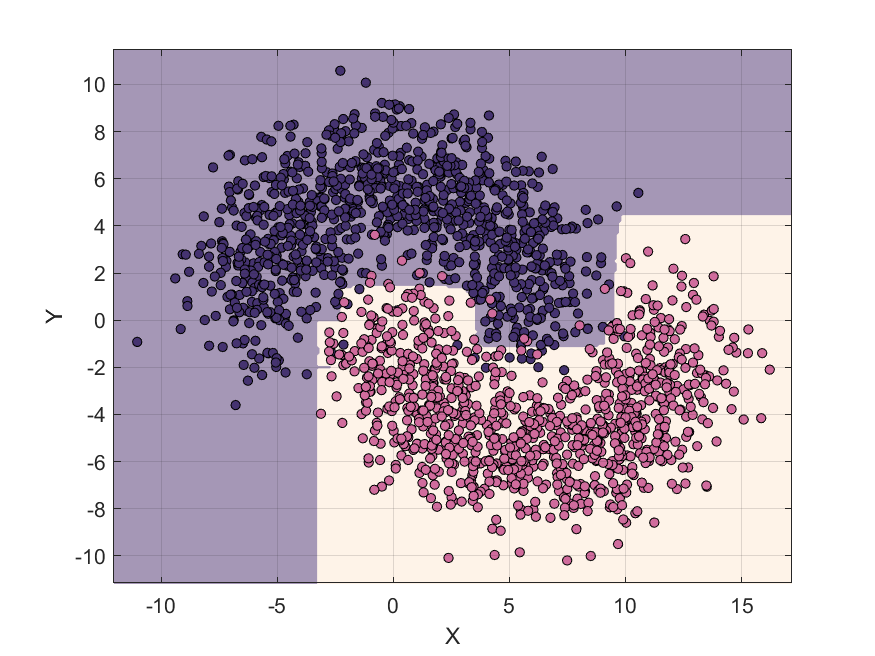}
	}
	% \hspace{-5mm}
        \subfigure[SVC]
	{
		\includegraphics[width=0.22\textwidth]{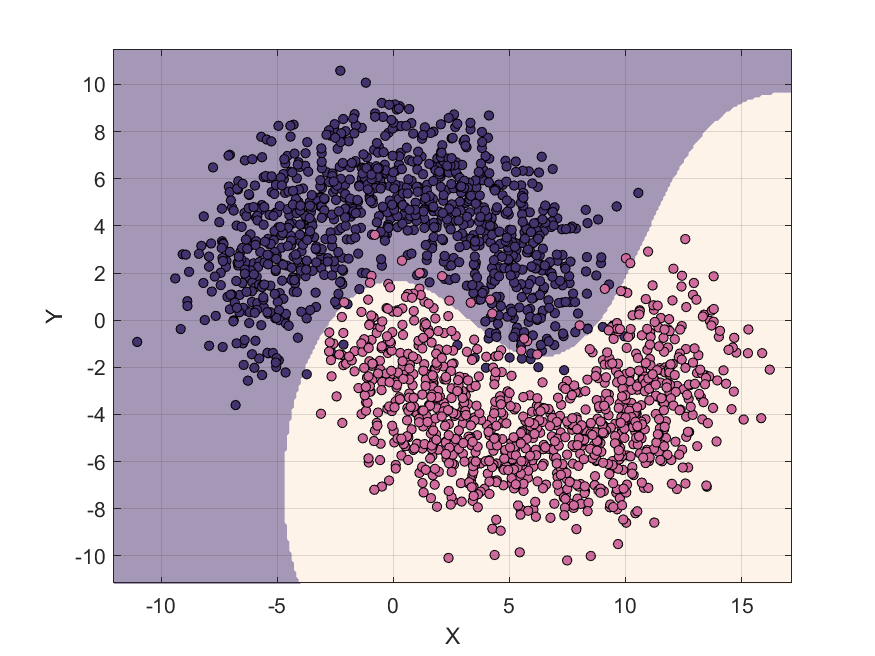}
	}
	% \hspace{-5mm}
         \subfigure[XGBOOST]
	{
		\includegraphics[width=0.22\textwidth]{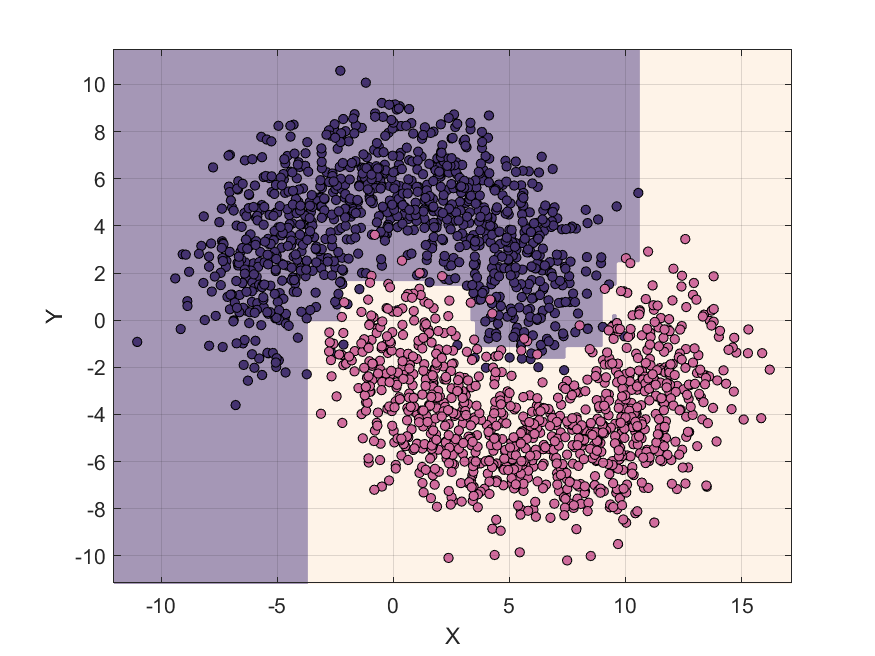}
	}
	% \hspace{-5mm}
        \subfigure[LIGHTGBM]
	{
		\includegraphics[width=0.22\textwidth]{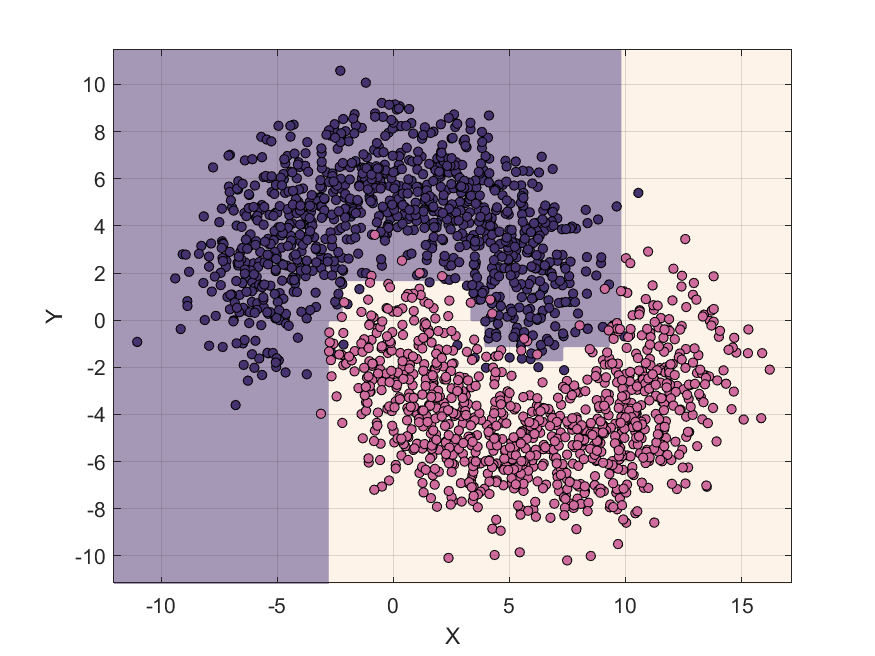}
	}
	% \hspace{-5mm}
	\caption{The decision boundaries of various algorithms on the toy dataset. (a) DATA. (b) OUR. (c) RF100. (d) SVC. (e) XGBOOST. (f) LIGHTGBM.}
	\label{F6}
\end{figure}
\subsection{Experimental Methodology}
To rigorously evaluate our proposed algorithm, we designed a comprehensive experimental framework utilizing a diverse dataset. This dataset comprises a moon-shaped toy dataset and 8 real-world datasets (BASEHOCK, breast uni, chess, iris, jaffe, pathbased,  RELATHE, and wine)\cite{yuan2024doubly ,yang2024fast ,shan2024random , chang2008robust , lyons1998coding}, encompassing a wide range of sample sizes and class numbers. We benchmarked our algorithm against Support Vector Classification (SVC), XGBoost, LightGBM and Random Forests with varying two numbers (50, 100), ensuring identical maximum depths for all tree-based models to maintain fairness. Our algorithm's hyperparameters were carefully tuned, with $\alpha$ fixed at 10 and $\gamma$ randomly selected from $\{5, 10, 15, 20, 25\}$ to test stability. Employing an $80:20$ train-test split, we evaluated performance using training set accuracy, test set accuracy, and overall accuracy to gain comprehensive insights into the algorithm's capabilities. \textbf{Our random seed is fixed at 1, and we are about to open-source the code.}
% 为严格评估我们提出的算法，我们设计了一个全面的实验框架，使用了多样化的数据集。该数据集包括一个……玩具数据集和10个真实世界数据集（TR41、warpPIE10P、Movement、Ecoil、Arcene、REUT、GLI85、Carcinom、Lung和Isolet），涵盖了广泛的样本规模和类别数。我们将我们的算法与支持向量分类（SVC）、XGBoost以及不同树数量（10、50、100）的随机森林进行了对比，确保所有基于树的模型使用相同的最大深度以保持公平性。我们算法的超参数经过精心调整，\alpha固定为10，\gamma 从{5, 10, 15, 20, 25}中随机选择以测试稳定性。采用80:20的训练集-测试集分割，我们通过训练集准确率、测试集准确率和总体准确率来评估性能，以全面了解算法的能力。

\subsection{Experimental Results and Analysis}
The experimental results reveal significant advantages of our algorithm across multiple dimensions. \textbf{Notably, despite utilizing only 10 decision trees, our algorithm consistently outperformed Random Forests with 100 trees on several datasets}, demonstrating its exceptional efficiency in handling large-scale data classification tasks. Fig.1 shows the decision boundary of our algorithm on the toy dataset, demonstrating high accuracy in plane partitioning. This "less is more" performance underscores the effectiveness of our fine-grained approach in optimizing computational resources while maintaining high accuracy.
% 实验结果揭示了我们算法在多个维度上的显著优势。值得注意的是，尽管仅使用10棵决策树，我们的算法在多个数据集上持续超越了使用100棵树的随机森林，展示了其在处理大规模数据分类任务时的卓越效率。这种"以少胜多"的表现凸显了我们细粒度方法在优化计算资源同时保持高准确率方面的有效性。（看看是否写以少胜多，想不到别的词了）

Our algorithm's generalization capability is particularly noteworthy. Although our algorithm does not achieve the best performance on every dataset, it surpasses the Random Forest algorithm using only $\frac{1}{10}$ of the trees, while maintaining optimal generalization ability.This pattern strongly indicates successful mitigation of overfitting, a crucial factor for model robustness in real-world applications.
% 我们算法的泛化能力尤为显著。通过分析训练集和测试集的准确率，我们发现虽然我们的算法在训练集上可能并不总是达到最高准确率，但在测试集和整体数据集上经常获得最佳性能。这种模式强烈表明成功抑制了过拟合，这对于实际应用中的模型鲁棒性至关重要。

%Visual analysis on the toy dataset further corroborated these findings, showing improved learner performance near classification boundaries. This precision in complex boundary regions elucidates the algorithm's excellent performance across various real-world datasets. Additionally, convergence studies demonstrated that our algorithm typically achieves convergence within 10 iterations, highlighting its computational efficiency and potential for processing large-scale datasets.
% 对玩具数据集的可视化分析进一步证实了这些发现，显示在分类边界附近学习器性能得到改善。这种在复杂边界区域的精确性阐明了算法在各种真实世界数据集上的出色表现。此外，收敛性研究表明我们的算法通常在10次迭代内实现收敛，突显了其计算效率和处理大规模数据集的潜力。

% In conclusion, our experimental results provide compelling evidence of the algorithm's superiority in balancing accuracy, computational efficiency, and generalization ability. By achieving fine-grained ensemble learning across diverse datasets, offering a powerful tool for addressing real-world classification challenges. Future research directions include exploring the algorithm's applicability to larger-scale, more complex datasets and extending this methodology to other machine learning paradigms.
% 总之，我们的实验结果为算法在平衡准确性、计算效率和泛化能力方面的优越性提供了令人信服的证据。通过在多种数据集上实现细粒度集成学习，我们的方法不仅验证了理论假设，还为机器学习研究开辟了新的途径。这种创新方法代表了在调和模型复杂性和预测性能方面的重大进展，为解决现实世界的分类挑战提供了强有力的工具。未来研究方向包括探索算法在更大规模、更复杂数据集上的应用性，以及将这种方法扩展到其他机器学习范式。（你看看要不要保留最后这句话，略有点扯淡，就是不知道实验discussion写不写这种）

%% file: tex/5_conclusion.tex
\section{CONCLUSION}
\label{sec:majhead}
In this paper, we introduce a novel fine-grained ensemble learning method achieving superior performance with fewer base learners. Our approach uses a learnable confidence matrix and a margin-based loss function, enabling precise optimization across categories and enhancing generalization. We provide theoretical guarantees on the loss function's local convexity and Lipschitz continuity. Experiments show our method outperforms random forests with 100 trees using only 10 base learners. The algorithm's rapid convergence and generalization highlight its potential for large-scale, resource-constrained applications. Future work will explore scalability to more complex datasets and adaptability to other machine learning paradigms.